\documentclass[sigconf]{acmart}

\usepackage{amsmath,amsfonts,bm}

\def\eqref#1{equation~\ref{#1}}
\def\1{\bm{1}}

\def\vh{{\bm{h}}}

\def\vx{{\bm{x}}}

\def\mA{{\bm{A}}}

\def\mM{{\bm{M}}}

\def\mX{{\bm{X}}}

\DeclareMathAlphabet{\mathsfit}{\encodingdefault}{\sfdefault}{m}{sl}
\SetMathAlphabet{\mathsfit}{bold}{\encodingdefault}{\sfdefault}{bx}{n}

\def\gE{{\mathcal{E}}}

\def\gG{{\mathcal{G}}}

\def\gO{{\mathcal{O}}}

\def\gV{{\mathcal{V}}}

\def\sP{{\mathbb{P}}}

\def\sR{{\mathbb{R}}}

\def\emA{{A}}

\DeclareMathOperator*{\argmin}{arg\,min}

\newcommand{\stitle}[1]{\vspace{1ex}\noindent{\bf #1}}

\usepackage{bbm}
\usepackage{bbold}
\usepackage{listings}
\usepackage{algorithm} 
\usepackage{algpseudocode} 
\usepackage{xspace}
\usepackage{subcaption}

\usepackage{amsthm}
\definecolor{codegreen}{rgb}{0,0.6,0}
\definecolor{codegray}{rgb}{0.5,0.5,0.5}
\definecolor{codepurple}{rgb}{0.58,0,0.82}
\definecolor{backcolour}{rgb}{0.95,0.95,0.92}

\newcommand{\datasetFont}{\text}
\newcommand{\ours}{\datasetFont{MixupExplainer}\xspace}
\newcommand{\bashapes}{\datasetFont{BA-Shapes}\xspace}
\newcommand{\bacom}{\datasetFont{BA-Community}\xspace}
\newcommand{\treec}{\datasetFont{Tree-Circles}\xspace}
\newcommand{\treeg}{\datasetFont{Tree-Grid}\xspace}
\newcommand{\bamo}{\datasetFont{BA-2motifs}\xspace}
\newcommand{\mutag}{\datasetFont{MUTAG}\xspace}
\newtheorem{theorem}{Theorem}
\newtheorem{problem}{Problem}

\newenvironment{proofs}{%
  \renewcommand{\proofname}{Proof Sketch}\proof}{\endproof}

\lstdefinestyle{mystyle}{
    backgroundcolor=\color{backcolour},   
    commentstyle=\color{codegreen},
    keywordstyle=\color{magenta},
    numberstyle=\tiny\color{codegray},
    stringstyle=\color{codepurple},
    basicstyle=\ttfamily\footnotesize,
    breakatwhitespace=false,         
    breaklines=true,                 
    captionpos=b,                    
    keepspaces=true,                 
    numbers=left,                    
    numbersep=5pt,                  
    showspaces=false,                
    showstringspaces=false,
    showtabs=false,                  
    tabsize=2
}

\AtBeginDocument{%
  \providecommand\BibTeX{{%
    \normalfont B\kern-0.5em{\scshape i\kern-0.25em b}\kern-0.8em\TeX}}}

\setcopyright{acmlicensed}
\acmConference[KDD '23]{Proceedings of the 29th ACM SIGKDD Conference on Knowledge Discovery and Data Mining}{August 6--10, 2023}{Long Beach, CA, USA}
\copyrightyear{2023}
\acmYear{2023}
\acmBooktitle{Proceedings of the 29th ACM SIGKDD Conference on Knowledge Discovery and Data Mining (KDD '23), August 6--10, 2023, Long Beach, CA, USA}
\acmPrice{15.00}
\acmISBN{979-8-4007-0103-0/23/08}
\acmDOI{10.1145/3580305.3599435}

\begin{document}

\title{\ours: Generalizing Explanations for Graph Neural Networks with Data Augmentation}

\author{Jiaxing Zhang}
\authornote{Both authors contributed equally to this research.}
\email{jz48@njit.edu}
\orcid{0009-0007-8031-661X}
\affiliation{%
  \institution{New Jersey Institute of Technology}
  \streetaddress{323 Dr Martin Luther King Jr Blvd}
  \city{Newark}
  \state{New Jersey}
  \country{USA}
  \postcode{07102-1982}
}

\author{Dongsheng Luo}
\email{dluo@fiu.edu}
\orcid{0000-0003-4192-0826}
\authornotemark[1]
\affiliation{%
  \institution{Florida International University }
  \streetaddress{11200 SW 8th Street}
  \city{Miami}
  \state{Florida}
  \country{USA}
  \postcode{33199}
}
\author{Hua Wei}
\authornote{Corresponding author}
\email{hua.wei@asu.edu}
\orcid{0000-0002-3735-1635}
\affiliation{%
  \institution{Arizona State University}
  \streetaddress{State University Ste 6}
  \city{Tempe}
  \state{Arizona}
  \country{USA}
  \postcode{85287}
}
\renewcommand{\shortauthors}{Jiaxing Zhang, Dongsheng Luo, \& Hua Wei}
\begin{abstract}
Graph Neural Networks (GNNs) have received increasing attention due to their ability to learn from graph-structured data. However, their predictions are often not interpretable. Post-hoc instance-level explanation methods have been proposed to understand GNN predictions. These methods seek to discover substructures that explain the prediction behavior of a trained GNN. In this paper, we shed light on the existence of the distribution shifting issue in existing methods, which affects explanation quality, particularly in applications on real-life datasets with tight decision boundaries. %While the OOD issue has been explored in the computer vision field, it has received less attention in the graph domain. 
To address this issue, we introduce a generalized Graph Information Bottleneck (GIB) form that includes a label-independent graph variable, which is equivalent to the vanilla GIB. Driven by the generalized GIB, we propose a graph mixup method, \ours, with a theoretical guarantee to resolve the distribution shifting issue. We conduct extensive experiments on both synthetic and real-world datasets to validate the effectiveness of our proposed mixup approach over existing approaches. We also provide a detailed analysis of how our proposed approach alleviates the distribution shifting issue.
\end{abstract}






\maketitle

\section{Introduction}
\label{sec:intro}
%\begin{itemize}
%\item Para. 1: Introduction to Graph and GNN
    
Graph Neural Networks (GNNs)~\cite{scarselli09gnnmodel}, a powerful technology for learning knowledge from graph-structured data, are gaining increasing attention in today's world, where graph-structured data such as social networks~\cite{fan19social, min21social}, molecular structures~\cite{chereda2019utilizing, mansimov19deepggnn}, traffic flows~\cite{wang20traffic, Li_Zhu_2021_traffic,wu19graphwave,ijcai2018p0505}, and knowledge graphs~\cite{sorokin-gurevych-2018-modeling-knowledgegraph} are widely used. GNNs work by propagating and fusing messages from neighboring nodes on the graph using message-passing mechanisms. These networks have achieved state-of-the-art performance in tasks like node classification, graph classification, graph regression, and link prediction.

Despite their success, GNNs, like other neural networks, lack interpretability. Understanding how GNNs make predictions is crucial for several reasons. First, it can increase user confidence when using GNNs in high-stakes applications~\cite{yuan2022explainability,longa2022explaining}. Second, it enhances the transparency of the models, making them suitable for use in sensitive fields such as healthcare and drug discovery, where fairness, privacy, and safety are critical concerns~\cite{zhang2022trustworthy,wu2022survey,li2022survey}. Thus, exploring the interpretability of GNNs is essential.

A common solution to improve GNN models' transparency is applying post-hoc instance-level explainability methods. These methods identify key substructures in input graphs to explain predictions made by trained GNN models, making it easier for humans to understand the models' inner workings. Examples of such methods include GNNExplainer~\cite{ying2019gnnexplainer}, which determines the importance of nodes and edges through perturbation, and PGExplainer\cite{luo2020parameterized}, which trains a graph generator to incorporate global information. Recent studies in the field~\cite{fang2023on, NEURIPS2021_be26abe7} also contribute to the development of these methods. Post-hoc explainability methods can be classified under a label-preserving framework, where the explanation is a substructure of the original graph and preserves the information about the predicted label. On top of the intuitive principle, Graph Information Bottleneck (GIB)~\cite{wu2020graph,miao2022interpretable,yu2020graph} maximizes the mutual information $I(G^{*}, Y)$ between the target label $Y$ and the explanation $G^{*}$ while constraining the size of the explanation as the mutual information between the original graph $G$ and the explanation $G^{*}$.

\begin{figure}
    \centering
    \includegraphics[width=0.48\textwidth]{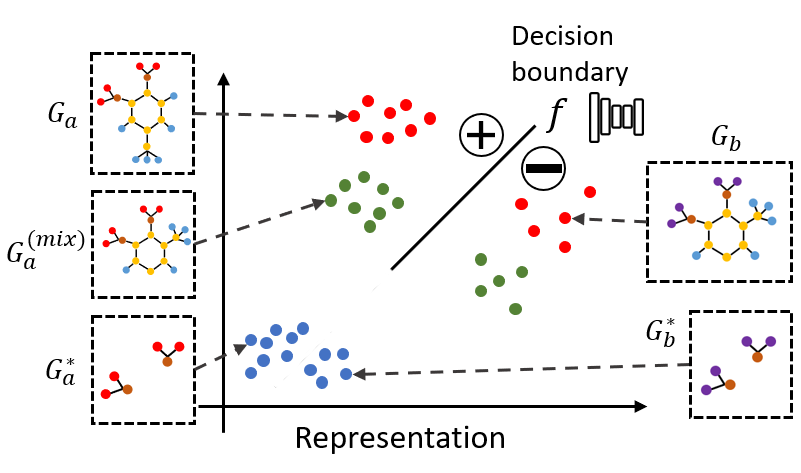}
    \vspace{-3mm}
    \caption{Visualization of original graphs $G$, explanation subgraphs $G^*$, and our generated graphs $G^{(\text{mix})}$. There is a large distributional divergence between explanation subgraphs $G^*$ and original graphs $G$. $G_a$ and $G_b$ are two graphs in the original dataset. More experimental results on the existence of the distributional divergence can be found in Section~\ref{sec:rq2}.}
    \vspace{-3mm}
    \label{fig:motivation}
\end{figure}

Approximating the mutual information between the label $Y$ and explanation $G^{*}$ is challenging due to its intractability, so previous works~\cite{ying2019gnnexplainer,luo2020parameterized,miao2022interpretable} usually estimate $I(G^{*}, Y )$ using $I(f(G^{*}), Y)$, the mutual information between the predictions $f(G^{*})$ from GNN model $f$ and its label $Y$. However, this approximation overlooks the distribution shifting issue between the original graph $G$ and explanation $G^{*}$ after the processing of the prediction model $f$. Due to differences in properties like the number of nodes or the structures in $G$, $G^{*}$ could have a different distribution from $G$. As seen in Figure~\ref{fig:motivation}, the visualization of the embeddings for the original graph and its explanation shows that the explanation embeddings are out of distribution with respect to the original graphs, which leads to impaired safe usage of the approximation because of the inductive bias in $f$. The negative impact of the distribution shifting problem on explanation quality is especially pronounced when applied to complex real-world datasets with tight decision boundaries.

%  Due to the xx, the GIB framework may lead to suboptimal explanations.

While the distribution shifting issue in post-hoc explanations has gained growing attention in computer vision~\cite{https://doi.org/10.48550/arxiv.1807.08024}, this issue is less explored in the graph domain. In computer vision, ~\cite{https://doi.org/10.48550/arxiv.1807.08024} optimizes image classifier explanations to highlight contextual information relevant to the prediction and consistent with the training distribution. ~\cite{10.1145/3485447.3512254} addresses the distribution shifting issue in image explanation via a module that quantifies affinity between perturbed data and original dataset distribution. In the graph domain, while a recent work~\cite{fang2023on} attempts to address distribution shifting by annealing the size constraint coefficient at the start of the explanation process, the distribution shifting issue still persists throughout the explanation process.

%  which means that there is still a difference between $I(f(\mathcal{G}^{*});\mathcal{Y})$ and $I(\mathcal{G}^{*};\mathcal{Y})$\cite{anonymous2023on}. So, the distribution shifting issue is still a challenge for GNNs' to explain. 
    
    %\item para 5: Our method: (1) An improvement to the previous Mixup approach (2) a novel objective function (3) Efficient instantiation with MixUp. Then summarize our contribution (maybe with items).
    
% To address the above challenge, we first derive a general form of GIB by including another label-irrelevant graph variable $G^\Delta$. We show that equivalent of our objective and the vanilla GIB \dongsheng{introduce GIB}. Moreover, with an appropriate $G^\Delta$, we can avoid the OOD problem in the following approximation with theoretical guarantees. Within the novel framework, we further propose a simple yet effective explanation method, MixupExplainer, with an improved Mixup approach. Specifically, with a safe assumption that a non-explainable part in a graph is label-irrelevant, we mix up the explanation with the non-explainable structure from another graph which is randomly sampled from the dataset. The explanation substructure is then obtained by minimizing the difference between the predicted labels of the original graph and the mixup graph. 

To address the distribution shifting issue in post-hoc graph explanation, we introduce a general form of Graph Information Bottleneck (GIB) that includes another label-independent graph variable $G^\Delta$. This new form of GIB is proven equivalent to vanilla GIB. By having $G^\Delta$ in the objective, we can alleviate the distribution shifting problem with theoretical guarantees. To further improve the explanation method, we propose MixupExplainer using an improved Mixup approach. The MixupExplainer assumes that a non-explainable part of a graph is label-independent and mixes the explanation with a non-explainable structure from another randomly sampled graph. The explanation substructure is obtained by minimizing the difference between the predicted labels of the original graph and the mixup graph.

% We first improved the traditional Mixup formula, making it able to mix the explanation with the base graph together. Instead of mixing a graph with another graph directly, we mix the explanation with the base structure from another graph randomly sampled from the dataset. Then we applied the mixup approach and explained the original graph with a new novel objection function to resolve the distribution shifting issue. By using $G^{*}+G^{\Delta}$, we could estimate the distribution of original graph $\mathcal{G}$ and minimize the difference of 
% $I(f(\mathcal{G}^{*}, \mathcal{G}^{\Delta}); \mathcal{Y})$ and $I(\mathcal{G};\mathcal{Y})$.
% We finally evaluated our approach on the six most commonly used datasets and achieved improvements in all of them. 
To the end, we summarize our contributions as follows.
    \begin{itemize}
    \item For the first time, we point out that the distribution shifting problem is prevalent in the most popular post-hoc explanation framework for graph neural networks. 
    \item We derive a generalized framework with a solid theoretical foundation to alleviate the problem and propose a straightforward yet effective instantiation based on mixing up the explanation with a randomly sampled base structure by aligning the graph and mixing the graph masks.
    \item Comprehensive empirical studies on both synthetic and real-life datasets demonstrate that our method can dramatically and consistently improve the quality of the explanations, with up to $35.5\%$ in AUC scores.
    \end{itemize}
% %\end{itemize}
% \begin{figure}
%     \centering
%     \includegraphics[width=0.48\textwidth]{figure/problem.png}
%     \caption{Problem Definition}
%     \label{fig:my_label}
% \end{figure}

\section{Related Work}
\label{sec:relatedwork}

\subsection{Graph Neural Networks}

The use of graph neural networks (GNNs) is on the rise for analyzing graph structure data, as seen in recent research studies~\cite{dai2022towards, fan19social, hamilton2017inductive}. There are two main types of GNNs: spectral-based approaches~\cite{bruna2013spectral, kipf2016semi, tang2019chebnet} and spatial-based approaches~\cite{atwood2016diffusion, duvenaud2015convolutional, xiao2021learning}. Despite the differences, message passing is a common framework for both, using pattern extraction and message interaction between layers to update node embeddings. However, GNNs are still considered a black box model with a hard-to-understand mechanism, particularly for graph data, which is harder to interpret compared to image data. To fully utilize GNNs, especially in high-risk applications, it is crucial to develop methods for understanding how they work.

% Despite their differences, most GNN variants can be summarized with the message-passing framework, which is composed of pattern extraction and interaction modeling within each layer [12]. Specifically, GNNs model messages from node representations. These messages are then propagated with various message-passing mechanisms to refine node representations, which are then utilized for downstream tasks [13, 40, 43]. Explorations are made by disentangling the propagation process [30, 33, 44] or utilizing external prototypes [16, 35]. Despite their success in network representation learning, GNNs are uninterpretable black box models. It is challenging to understand their behaviors even if the adopted message-passing mechanism and parameters are given. Besides, unlike traditional deep neural networks where instances are identically and independently distributed, GNNs consider node features and graph topology jointly, making the interpretability problem more challenging to handle.

\subsection{GNN Explanation}

Many attempts have been made to interpret GNN models and explain their predictions~\cite{rgexp, ying2019gnnexplainer, luo2020parameterized, subgraphx, spinelli2022meta, wang2021causal}. These methods can be grouped into two categories based on granularity: (1) instance-level explanation, which explains the prediction for each instance by identifying significant substructures~\cite{ying2019gnnexplainer, subgraphx, rgexp}, and (2) model-level explanation, which seeks to understand the global decision rules captured by the GNN~\cite{luo2020parameterized, spinelli2022meta, bald19expgcn}. From a methodological perspective, existing methods can be classified as (1) self-explainable GNNs~\cite{bald19expgcn, dai21towards}, where the GNN can provide both predictions and explanations, and (2) post-hoc explanations~\cite{ying2019gnnexplainer, luo2020parameterized, subgraphx}, which use another model or strategy to explain the target GNN. In this work, we focus on post-hoc instance-level explanations, which involve identifying instance-wise critical substructures to explain the prediction. Various strategies have been explored, including gradient signals, perturbed predictions, and decomposition.

Perturbed prediction-based methods are the most widely used in post-hoc instance-level explanations. The idea is to learn a perturbation mask that filters out non-important connections and identifies dominant substructures while preserving the original predictions. For example, GNNExplainer~\cite{ying2019gnnexplainer} uses end-to-end learned soft masks on node attributes and graph structure, while PGExplainer~\cite{luo2020parameterized} incorporates a graph generator to incorporate global information. RG-Explainer~\cite{rgexp} uses reinforcement learning technology with starting point selection to find important substructures for the explanation.

However, most of these methods fail to consider the distribution shifting issue. The explanation should contain the same information that contributes to the prediction, but the GNN is trained on a data pattern that consists of an explanation subgraph relevant to labels, and a label-independent structure, leading to a distribution shifting problem when feeding the explanation directly into the GNN. Our method aims to capture the distribution information of the graph and build the explanation with a label-independent structure to help the explainer better minimize the objective function and retrieve a higher-quality explanation.

\subsection{Graph Data Augmentation with Mixup}
Data augmentation addresses issues such as noise, scarcity, and out-of-distribution problems. One popular data augmentation approach is using Mixup~\cite{mixupon17} strategy to generate synthetic training examples based on feature mixing and label mixing. Specifically,~\cite{verma2021graphmix,wang2021mixup} mix the graph representation learned from GNNs to avoid dealing with the arbitrary structure in the input space for mixing a node or graph pair. ifMixup~\cite{guo2021ifmixup} interpolates both the node features and the edges of the input pair based on feature mixing
and graph generation. ~\cite{gmixup} and ~\cite{wu2021graphmixup} generate interpolated graphs with the estimation of the properties in the graph data, like the graphon of each class or nearest neighbors of target nodes. All the previous methods~\cite{verma2019manifold,verma2021graphmix,gmixup,wang2021mixup,guo2021ifmixup} aim to generalize the mixup approach to improve the performance of classification models like GNNs. Unlike existing graph mixup approaches, this paper solves a different task, which is to generalize the explanations for GNN.

% Graph augmentation addresses issues such as graph noise, scarcity, and out-of-distribution (OOD) problems. One popular approach for data augmentation is Mixup. Some works have further developed Mixup to solve OOD problems. For example, G-Mixup\cite{gmixup} explores the graphon feature of a dataset, mixing different graph structures by mixing the graphon matrices to generate a mixed graph that contains elements from both types. C-Mixup\cite{cmix} is another way of Mixup for improving generalization, which mixes images of objects with different rotations in space, generating a middle-pose image to avoid OOD problems. These works both solve OOD issues in specific domains and improve GNN performance. Inspired by these works, we aim to mix explanations and base structures to address the distribution shifting issue during graph explanation.~\hua{C-mixup is not graph augmentation, need to rewrite this paragraph} \jx{revised a bit}
\section{Preliminary}

\subsection{Notations and Problem Definition}
% \subsubsection{Introduce the method}
We denote a graph as $G= (\mathcal{V}, \mathcal{E}; \mX, \mA)$, where $\mathcal{V} = \{v_1, v_2, ..., v_n\}$ represents a set of $n$ nodes and $ \mathcal{E} \in \mathcal{V} \times \mathcal{V}$ represents the edge set. Each graph has a feature matrix $ \mX \in \sR^{n\times d} $ for the nodes, where in $\mX $, $ \vx_i  \in \sR^{1\times d} $ is the $d$-dimensional node feature of node $v_i$. $\mathcal{E}$ is described by an adjacency matrix $ \mA \in \{0,1\}^{n\times n}$. $\emA_{ij} = 1$ means that there is an edge between node $v_i$ and $v_j$; otherwise, $\emA_{ij} = 0$. 

For graph classification task, each graph $G_i$ has a label $Y_i \in \mathcal{C}$, with a GNN model $f$ trained to classify $G_i$ into its class, i.e., $f:(\mX, \mA) \mapsto \{1, 2, ..., C\}$. 
For the node classification task, each graph $G_i$ denotes a $K$-hop sub-graph centered around node $v_i$, with a GNN model $f$ trained to predict the label for node $v_i$ based on the node representation of $v_i$ learned from $G_i$.

\begin{problem} [Post-hoc Instance-level GNN Explanation]
\label{prob:exp}
Given a trained GNN model $f$, for an arbitrary input graph $G= (\mathcal{V}, \mathcal{E}; \mX, \mA)$, the goal of post-hoc instance-level GNN explanation is to find a subgraph $G^{*}$ that can explain the prediction of $f$ on $G$. 
\end{problem}
Informative feature selection has been well studied in non-graph structured data~\cite{Li17featureselect},  and traditional methods, such as concrete autoencoder~\cite{balin2019concrete}, can be directly extended to explain features in GNNs. In this paper, we focus on discovering important typologies. Formally, the obtained explanation $G^{*}$ is depicted by a binary mask $\mM \in \{0, 1\}^{n\times n}$ on the adjacency matrix, e.g., $G^{*} = ( \gV, \gE, \mA\odot \mM; \mX)$, $\odot$ means elements-wise multiplication. The mask highlights components of $G$ which are essential for $f$ to make the prediction. 
% \subsubsection{GNN Explanation}

% We denote an explanation as an edge mask $M_i$ for graph $G_i$, where $M_i = [w_1, w_2, ..., w_e]$, $e$ means the number if the edges in graph $G_i$.

\subsection{Graph Information Bottleneck}

\begin{figure}
\centering
    \begin{tabular}{cc}
    \includegraphics[height=0.345\linewidth]{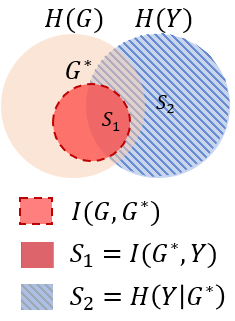}
        & \includegraphics[height=0.345\linewidth]{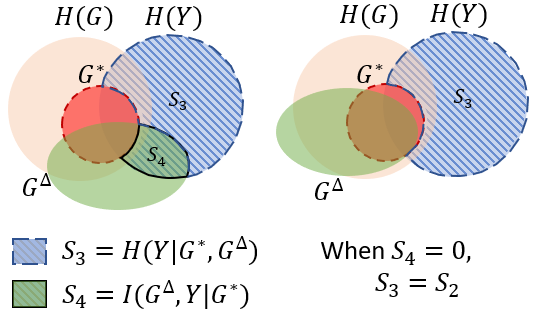} \\
        \small{
        \begin{tabular}[c]{@{}c@{}}(a) Previous GIB Objective\end{tabular}}& 
        \small{
        \begin{tabular}[c]{@{}c@{}}(b) Our Generalized GIB Objective\end{tabular}}
    \end{tabular} \\
    \caption{Illustration of GIB and our proposed new objective. (a) Previous vanilla GIB objective aims to minimize $I(G^{*},Y)$ and $H(Y|G^{*})$, with a smaller overlap between $G^{*}$ and $G$. (b) Our generalized GIB objective has the same objective as vanilla GIB, with a larger lap between $G$ and $G^{*}+G^\Delta$, resulting in less distribution shifting issue.}
    \label{fig:venn}
\end{figure}

% Note that: different from the introduction, we focus on the formal definition in this part. 
The Information Bottleneck (IB)~\cite{tishby2000information,tishby2015deep} provides an intuitive principle for learning dense representations that an optimal representation should contain \textit{minimal} and \textit{sufficient} information for the downstream prediction task. Based on IB, a recent work unifies the most existing post-hoc explanation methods for GNN, such as GNNExplainer~\cite{ying2019gnnexplainer}, PGExplainer~\cite{luo2020parameterized}, with the graph information bottleneck (GIB) principle~\cite{wu2020graph,miao2022interpretable,yu2020graph}. Formally, the objective of  explaining the prediction of $f$ on $G$ can be represented by 
% Finding subgraphs that can the  (Wu et al., 2020; Yu et al., 2021). 
% The objective explanation models for GNN can be formulated as the Graph Information Bottleneck (GIB) objective~\cite{ying2019gnnexplainer,luo2020parameterized,miao2022interpretable}.  
% \begin{equation}
%     \argmax{G_e}{ I(G_e, y)} = H(y)-H(y|G_e),
% \end{equation}
\begin{equation}
    \label{eq:gib}
    \argmin_{G^*} I(G, G^*)-\alpha I(G^*,Y),
\end{equation}
where $G^*$ is the explanation subgraph, $Y$ is the original or ground truth label, and $\alpha$ is a hyper-parameter to get the trade-off between minimal and sufficient constraints.
GIB uses the mutual information $I(G, G^*)$ to select the minimal explanation that inherits only the most indicative information from $G$ to predict the label $Y$ by maximizing $I(G^*, Y)$, where $I(G, G^*)$ avoids imposing potentially biased constraints, such as the size or the connectivity of the selected subgraphs~\cite{miao2022interpretable}. Through the optimization of the subgraph, $G^*$ provides model interpretation. Further, from the definition of mutual information, we have $I(G^*, Y) = H(Y)-H(Y|G^*)$, where the entropy $H(Y)$ is static and independent of the explanation process. Thus, minimizing the mutual information between the explanation subgraph $G^*$ and $Y$ can be reformulated as maximizing the conditional entropy of $Y$ given $G^*$. Formally, we rewrite the GIB objective as follows: 
\begin{equation}
    \label{eq:gibp}
    \argmin_{G^*} {I(G, G^*)+\alpha H(Y|G^*)},
\end{equation}
As is shown in Figure~\ref{fig:venn}(a), the objective function in Eq.~(\ref{eq:gibp}) optimizes $G^*$ to have the minimal mutual information with the original graph $G$, which could be expressed as a subgraph from $G$ with a smaller size, or scattered components in $G$, while at the same time provides maximum mutual information for $Y$, which is equivalent to have minimum entropy $H(Y|G^*)$.

Due to the intractability of entropy of the label conditioned on explanation, a widely-adopted approximation in previous methods~\cite{ying2019gnnexplainer,luo2020parameterized,zhao2022consistency} is:
\begin{equation}
    \label{eq:gibp-ce}
    \argmin_{G^*} {I(G, G^*)+\alpha H(Y|G^*)} \approx \argmin_{G^*} {I(G, G^*)+\alpha CE(Y,Y^*)},
\end{equation}
where $Y^*=f(G^*)$ is the predicted label of $G^*$ made by the model to be explained, $f$ and the cross-entropy $\text{CE}(Y,Y^*)$ between the ground truth label $Y$ and $Y^*$ is used to approximate $H(Y|G^*)$. 
\section{Methodology}
\label{sec:method}

In this section, we first introduce an overlooked problem in the GIB objective. Then we propose a generalized GIB objective to address the problem, which directly inspires our method through a mixup approach.

% \subsection{A Novel Objective function}
% \begin{itemize}
%     \item issues with GIB (MMI) objective
%     \item introduces our objective. (Maybe with Figures)
% \end{itemize}

% \begin{figure*}
%     \centering
%     \begin{subfigure}[b]{0.5\textwidth}
%          \includegraphics[width=0.34\textwidth]{figure/GIB.png}
%          \subcaption{W/O Data Augmentation}
%          \label{fig:ablation:embedding:tsne_0}
%      \end{subfigure}
%     \hspace{-3em}
%     \begin{subfigure}[b]{0.5\textwidth}
%          \includegraphics[width=0.54\textwidth]{figure/ourframe.png}
%          \subcaption{W/ Data Augmentation}
%          \label{fig:ablation:embedding:tsne_02}
%      \end{subfigure}
%     \parbox{0.48\textwidth}{
%     \caption{T-SNE embedding visualization of contrastive learning pretrain model without/with 0.2 corruption data augmentation}
%         \label{fig:framework}
%     }
%     % \vspace{-3ex}
% \end{figure*}
\begin{figure*}
    \centering
    \includegraphics[width=0.95\textwidth]{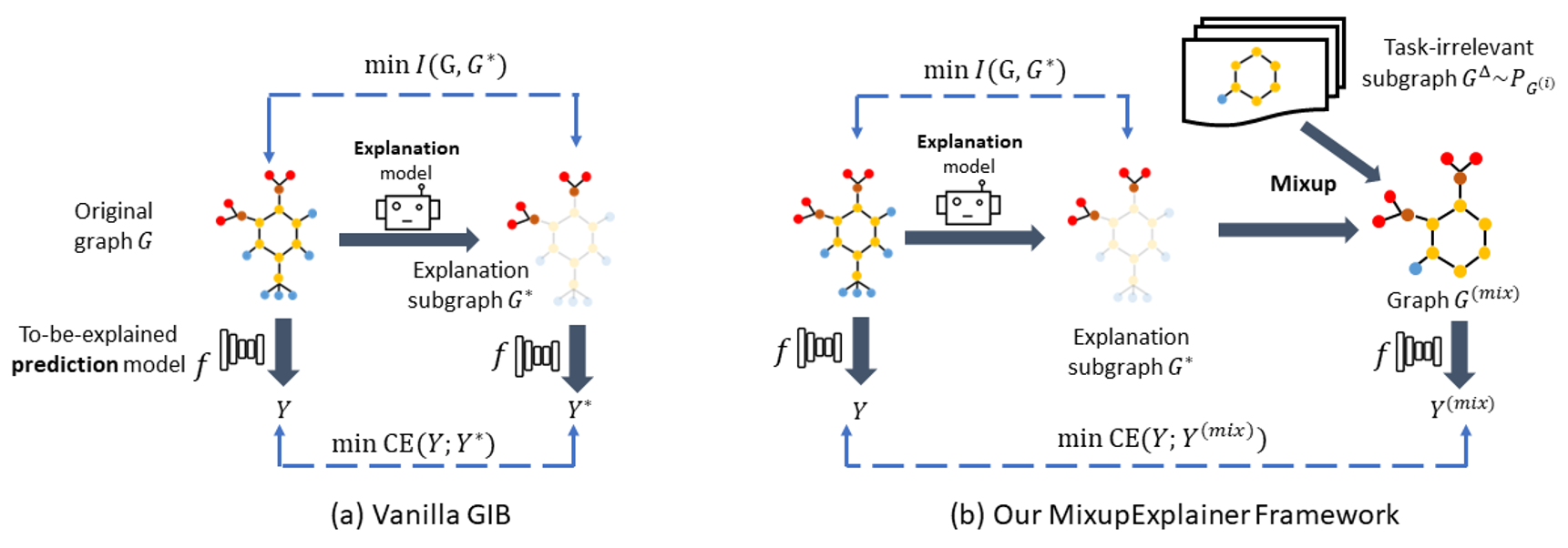}
    \caption{ Illustration of the GIB-based explanation and  our proposed \ours. (a) Vanilla GIB directly minimizes $\text{CE}(Y, Y^*)$, which is the cross entropy between the original prediction $Y$ and the prediction of explanation subgraph $G^*$ made by the to-be-explained model $f$. (b) Our \ours first generates an augmented graph $G^{(\text{mix})}$ by mixing up the explanation subgraph $G^*$ with the label-independent part from another randomly sampled graph. Then we minimize the cross entropy between $Y$ and $Y^{\text{(mix)}}$, the prediction made by $f$ on $G^{(\text{mix})}$.}
    \label{fig:framework}
\end{figure*}

\subsection{Generalized GIB}
\subsubsection{Diverging Distributions in Eq.~(\ref{eq:gibp-ce})}
Although prevalent, the approximation with $Y^*=f(G^*)$ in Eq.~(\ref{eq:gibp-ce}) overlooks the distributional divergence between the original graph $G$ and the dense subgraph $G^*$ after the processing of the prediction model $f$. An intuitive example from the \mutag dataset~\cite{debnath1991structure} is shown in Figure~\ref{fig:motivation}. The prediction model $f$, represented by a hypothesis line, performs well in classifying the positive and negative samples. Due to the distribution shifting problem naturally inherent in $f(G)$ and $f(G^*)$ on explanation subgraphs, $f$ maps some explanation subgraphs across the decision boundary to the negative region. As a result, the explanation subgraph achieved by Eq.~(\ref{eq:gibp-ce}) may be suboptimal and even far away from the ground truth explanation due to the significant divergence between $f(G^*)$ and $f(G)$. The existing GIB framework could work for simple synthetic datasets by relying on the implicit knowledge associated with the class and assuming a large decision margin between the two or more classes. However, in more practical scenarios like \mutag, the existing approximation may be heavily affected by the distribution shifting problem~\cite{miao2022interpretable, fang2023on}. 

%The distribution shift from $G$ to $G^*$ leads to diverging explanations as the prediction of $f$ needs not only the label-dependent subgraph $G^*$ but also the label-independent subgraph in $G$ due to information aggregation in $f$. The label-independent subgraph in $G_a$ with label $a$ also contains information. For example, connecting it with the label-independent subgraph will not lead to another label.

% The existing framework works well for simple synthetic datasets due to the large margins between decision boundaries and data samples. However, in more practical scenarios, the existing approximation may be heavily affected by the OOD problem~\cite{miao2022interpretable,anonymous2023on}. 

\subsubsection{Addressing with Label-independent Subgraph} To address the above challenge in the previous GIB methods, %which neglect the label-independent part of the original graph,
we first generalize the existing GIB framework by taking a label-independent subgraph $G^\Delta$ into consideration. The intuition is that for an original graph $G_a$ with label $Y_a$, the label-independent subgraph $G_a^\Delta$ also contains useful information. For example, $G_a^\Delta$ makes sure that connecting it with the label-preserving subgraph $G_a^*$ will not lead to another label. Formally, given a graph variable $G^\Delta$ that satisfies $I(G^\Delta, Y | G^*)=0$, the GIB objective can be generalized as follows.
\begin{equation}
    \label{eq:ggib}
    \argmin_{G^*}{I(G, G^*)+\alpha H(Y|G^*,G^\Delta)}, \quad \text{s.t. } \quad  I(G^\Delta,Y | G^*)=0.
\end{equation}

As shown below, our generalized GIB has the following property.

\stitle{Property 1.} \emph{ The generalized GIB objective, Eq. (\ref{eq:ggib}) is equivalent to vanilla GIB, Eq. (\ref{eq:gibp}).}

This can be proved by the definition of conditional entropy. With the condition that $I(G^\Delta,Y | G^*) =0$, we have $H(Y|G^*) = H(Y|G^*) + I(G^\Delta,Y | G^*)= H(Y|G^*, G^\Delta)$. Thus, the optimal solutions of GIB and our generalized version are equivalent. In addition, the advantage of our objective is that by choosing a suitable $G^\Delta$ that minimizes the distribution distance, $D(G^*+G^\Delta, G)$, we can approximate the GIB without including the distribution shifting problem. An intuitive illustration is given in Figure~\ref{fig:venn}(b).

Following exiting work~\cite{ying2019gnnexplainer,luo2020parameterized}, we can further approximate $H(Y|G^*,G^\Delta)$ with $\text{CE}(Y,Y^m)$, where $Y^m = f(G^*+G^\Delta)$ is the predicted label of $G^*+G^\Delta$ made by the model $f$ to be explained. Especially when $G^\Delta$ is an empty graph, our objective degenerates to the vanilla approximation.  Formally, we derive our new objective for GNN explanation as follows:

\begin{equation}
\begin{aligned}
\label{eq:gib-ours}
     \argmin_{G^\Delta, G^*} & { \quad I(G, G^*)+\alpha \text{CE}(Y, Y^m)}\\
    & \text{s.t. } \text{D}(G^*+G^\Delta, G) =0 , I(G^\Delta,Y | G^*)=0.
\end{aligned}
\end{equation}

% \begin{equation}
%     \argmax{G_e}{I(y,G_o)} \text{and } \argmin{G_e} KL(G_o, G) ,
% \end{equation}

% \begin{figure*}
%     \centering
%     \includegraphics[width=\textwidth]{figure/overview.png}
%     \caption{Our approach overview}
%     \label{fig:overview}
% \end{figure*}

% \begin{figure}
%     \centering
%     \includegraphics[width=0.4\textwidth]{figure/method.png}
%     \caption{Caption}
%     \label{fig:my_label}
% \end{figure}

% \begin{figure}
%     \centering
%     \includegraphics[width=0.4\textwidth]{figure/fig_mixup.png}
%     \caption{The overview of Explanation Mixup}
%     \label{fig:mixup}
% \end{figure}

\subsection{\ours}
Inspired by Eq.~\ref{eq:gib-ours}, in this section, we introduce a straightforward yet theoretically guaranteed instantiation, \ours, to resolve the distribution shifting issue. Figure~\ref{fig:framework} demonstrates the overall framework of the proposed \ours and the differences between \ours and previous GIB methods. \ours includes a graph generation phase after extracting the explanation of the graph with the explainer. Specifically, we instantiate the $G^\Delta$ from the distribution of label-independent subgraphs from the graph dataset, denoted as $\sP_{\gG^{(i)}}$, and connect $G^*$ and $G^\Delta$ to generate a new graph $G^{(\text{mix})}$. Formally,
\begin{equation}
\label{eq:mixupformula}
 G^\Delta \sim \sP_{\mathcal{G}^{(i)}}, \quad   G^{(\text{mix})} = G^* + G^\Delta.
\end{equation}
To avoid the trivial case that $G=G^{(\text{mix})}$, when sampling $G^\Delta$, we dismiss the original graph itself. In addition, since $G^\Delta$ is sampled without considering the label information, we can make a safe assumption that $I(G^\Delta, Y | G^*)=0$. 
% Next, we show that the mixup operation can preserve the graph distribution. 
% \subsubsection{Introduce the method}

As stated in Problem~\ref{prob:exp}, given a graph $G_a = (\mA_a, \mX_a)$\footnote{We dismiss $\gV$ and $\gE$ to simplify the notations.} and a to-be-explained model $f$, an explanation model $g$ aims to learns a subgraph $G^*_a$, represented with the edge mask $\mM_a=g(G_a)$ on the adjacency matrix $\mA_a$. To generate a graph distributed similarly to $G_a$, we need to generate a label-independent subgraph, where we randomly sample another graph instance from the dataset, denoted by $G_b$, without considering the label information. With the explanation model $g$, we obtain the corresponding edge mask $\mM_b$ for $G_b$.  Then, we mix these two graphs by connecting the informative part in $G_a$ and the label-independent part in $G_b$. 
We first assume that $G_a$ and $G_b$ share the same set of nodes, and more general cases are discussed in the next section. Formally, the mask of the mixed graph, $\mM_a^{(\text{mix})}$, is calculated as follows.
 % In our work, we mix up the explanation with a graph, which is sampled randomly from the data set, to maintain the graph distribution close to the original one. As shown in Fig.?., we generate the explanation $G^{*}_{a}$ and $G^{*}_{b}$ for the original graph $G_a$ and $G_b$ respectively, where the explanations are represented as Edge mask $\mM_a$ and $\mM_b$. We then mix up $G^{\prime}_{a}$ with the base graph of $G_b$ with this formula: 
 % A formal mathematical expression of Mixup is $X_{mix} = \lambda X_i + (1 - \lambda)x_j, y_{mix} = \lambda y_i + (1 - \lambda)y_j$, while $(x_i, y_i)$ and $(x_j, y_j)$ represent two samples of data and label respectively.
 \begin{equation}
 \label{eq:mixup-ours-lmd}
\mM_a^{(\text{mix})} = \lambda \mM_a + (\mA_b-\lambda \mM_b),
 \end{equation}
where $\mA_b$ is the adjacency matrix of graph $G_b$ and $\lambda$ is a hyper-parameter to support flexible usage of mixup operation. Then, we have $G^{(\text{mix})}_a = (\mX_a, \mM_a^{(\text{mix})})$.
The mask matrix $\mM_a$ and $\mM_b$ denote the weight of the edges in $\mA_a$ and $\mA_b$, respectively, with the same size of the matrix. By default, we mix up $G^{*}_{a}$ with the rest part of the $G_b$ by setting $\lambda = 1$ and above formula could be further simplified as:
 \begin{equation}
 \label{eq:mitakeours}
\mM_a^{(\text{mix})} = \mM_a + (\mA_b - \mM_b).
 \end{equation}

Note that our proposed mixup approach is different from traditional mixup approaches ~\cite{mixupon17, gmixup, yao2022c} in data augmentation, where they usually follow a form similar to $\mM^{(\text{mix})} = \lambda \mM_a + (1-\lambda)\mM_b$. This form of mixup does not differentiate label-dependent from label-independent parts. On the contrary, our proposed mixup approach in Eq.~(\ref{eq:mixup-ours-lmd}) includes the label-dependent part in $G_a$ with $\lambda \mM_a$ and excludes the label-dependent part in $G_b$ by subtracting the same $\lambda$ on $\mM_b$ from $\mA_b$.

% \stitle{difference}
% between our and traditional mixup

\subsubsection{Implementation} In this section, we introduce the implementation details of the mixup function and provide the pseudo-code of graph mixup in Algorithm~\ref{alg1}.

Given a graph $G_a$ with $n_a$ nodes and another graph $G_b$ with $n_b$ nodes, the addition in Eq.~(\ref{eq:mixup-ours-lmd}) between two matrices requires $\mM_a$ and $\mM_b$ have the same dimensions, i.e., $G_a$ and $G_b$ have the same number of nodes. However, in real-world graph datasets, this assumption may not hold, leading to a mismatch between the dimensions of $\mM_a$ and $\mM_b$.
In order to merge two graphs with different sets of nodes, we first extend node sets in $G_a$ and $G_b$ to a single node set $\gV_a \cup \gV_b$, and their adjacency matrices are calculated with the following functions:
\begin{equation}
\label{eq:align-adj-a}
    \mA_a^{\text{ext}} = \left[ \begin{array}{cc}
\mA_a & \mathbb{0} \\ \mathbb{0} & \mathbb{0}_b \end{array} \right],  \mA_b^{\text{ext}} = \left[ \begin{array}{cc}
\mathbb{0}_a & \mathbb{0} \\ \mathbb{0} & \mA_b \end{array} \right],
\end{equation}
where $\mathbb{0}_a$ and $\mathbb{0}_b$ are zero matrices with shapes  $n_a\times n_a$ and $n_b\times n_b$, respectively.

After extending $G_a$ and $G_b$, we then merge them into $G^{(\text{mix})} = (\mX^{(\text{mix})}, \mM_a^{\text(mix)}\odot \mA^{(\text{mix})})$, where $\mX^{(\text{mix})} = [\mX_a;\mX_b]$ is the concatenation of node features $\mX_a$ and $\mX_b$; $\mA^{(\text{mix})}$ is the merged adjacency matrix; $\mM_a^{\text(mix)}$ is the edge mask indicating the edge weights for the explanation. 

Specifically, the adjacency matrix of $G^{(\text{mix})}$ is:
% This process could be shown as the following equation:
\begin{equation}
\label{eq:align-adj}
    \mA^{(\text{mix})} = \left[ \begin{array}{cc}
\mA_a & \mA_c \\ \mA^T_c & \mA_b \end{array} \right],
\end{equation}
where $\mA_c$ is a matrix indicating the cross-graph connectivity between the nodes in $G_a$ and $G_b$. In practice, we randomly sample $\eta$ cross-graph edges to connect $G_a$ and $G_b$ at each mixup step to ensure the mixed graph is a connected graph to be optimized together on both label-dependent and label-independent subgraphs.
% 
% 
% different graphs when mixing them up. Therefore, we need to align the original graph with the randomly sampled graph from the dataset when applying our mixup method. 
% 
% In practice, 
% The way we take is connecting the explanation with the base structure with $\eta$ randomly sampled edges.~\hua{what does the previous sentence mean? Add more explanations} \jx{explained with the following instance}For instance, given a graph $G_a$ with $N_a$ nodes and another graph $G_b$ with $N_b$ nodes before we mix up $G_a$ with $G_b$, we would shift the nodes in $G_b$ by $n_i \leftarrow n_i + N_a$, where $n_i$ means the id for $ I_{th}$ node in graph $G_b$. Then we would merge the graph $G_a$ and $G_b$ together into $G_{mix}$ with size $N_a + N_b$. This process could be shown as the following equation:
% \begin{equation}
% \label{eq:align-adj}
%     \mA^{(\text{mix})} = \left[ \begin{array}{cc}
% \mA_a & \mA_c \\ \mA^T_c & \mA_b \end{array} \right]
% \end{equation}

% After aligning and merging the two graphs, we would sample $\eta$ edges between $G_a$ and $G_b$ as the connected matrix $\mA_c$, where the size of $A_c$ is $N_a \times N_b$. Then we would concatenate the $\mA_{mix}$ from $G_{mix}$ with $\mA_c$ together. The size of adjacency matrix $\mA_{mix}$ is $(N_a+N_b) \times (N_a+N_b)$.
Similarly, the edge mask matrix is obtained from extended $\mM_a$ and $\mM_b$ and calculated with Eq.~(\ref{eq:mixup-ours-lmd}). Formally, we have
\begin{equation}
\label{eq:align-mask}
    \mM^{(\text{mix})}_a = \left[ \begin{array}{cc}
\lambda \mM_a & \mM_c \\ \mM^T_c & 
\mA_b-\lambda \mM_b \end{array} \right]
\end{equation}
where the explainer $ g$ gives the $\mM_a$ and $\mM_b$, $\mM_c$ is the weight matrix on the randomly-sampled cross-graph edges corresponding with $\mA_c$, where the values are randomly sampled on connected edges in $\mA_c$ at each mixup step and thus will not be optimized by $g$.
 
Finally, we can mixup the edge weight matrices $\mM_a^{\text{ext}}$ and $\mM_b^{\text{ext}}$ together with Eq.~(\ref{eq:mixup-ours-lmd}). 
% The mixed graph $G^{(\text{mix})}$ will have a mixed feature matrix $\mX^{(\text{mix})}$ concatenated directly from $\mX_a$ and $\mX_b$, and mixed edge weight $\mM^{(\text{mix})}$, which
The mixed graph $G_a^{(\text{mix})}$ is then fed into the GNN model $f$ to calculate the predicted result $Y^{(\text{mix})}$. 
The detailed implementation is shown in Algorithm \ref{alg1}.
%  shows how we do the graph alignment and mixup.

\subsubsection{Computational Complexity Analysis} Here, we analyze the computational complexity of our mixup approach. 
% The major computation costs come from the extension of two graphs to match their dimensions. 
Given a graph $G_a$ and a randomly sampled graph $G_b$, the complexity of graph extension on adjacency matrices and edge masks is $\gO(|\gE_a|+|\gE_b|)$, where $|\gE_a|$ and $|\gE_b|$ denote the number of edges in $G_a$ and $G_b$, respectively. To generate $\eta$ cross-graph edges, the computational complexity is $\gO(\eta)$. For mixup, the complexity is $\gO(|\gE_a|+|\gE_b|)$. By considering $\eta$ as a small constant, the overall complexity of our mixup approach is $\gO(|\gE_a|+|\gE_b|)$.

\subsubsection{Theoretical Justification}
In the following, we theoretically prove that: \emph{the proposed mixup approach could reduce the distance between the explanation and original graphs.} Formally, we have the following theorem:

\begin{theorem}
Given an original graph $G$, graph explanation $G^*$ and $G^{(\text{mix})}$ generated by Eq.~(\ref{eq:mixup-ours-lmd}), we have $KL(G, G^*) \geq \text{KL}(G, G^{(\text{mix})})$.
\end{theorem}

\begin{proofs} 
% The original form of Mixup approach: $$G^{(\text{mix})} = \lambda \mM_a + (1-\lambda)\mM_b$$
% Our mixup form: $$G^{(\text{mix})} = \lambda \mM_a + (\mathbbm{1}-\lambda \mM_b)$$
According to the previous work~\cite{ying2019gnnexplainer, luo2020parameterized}, a graph $G$ can be 
treated as $G = G^{(e)} + G^{(i)}$, where $G^{(e)}$ presents the underlying subgraph that makes important contributions to GNN’s predictions, which is the expected explanatory graph, and $G^{(i)}$ consists of the remaining label-independent edges for predictions made by the GNN. 
%For example, in data-set BA-2motifs, a graph consists of a BA random graph as a base and a motif (a house or a circle) as ground truth. The GNN model is trained to classify the graph into two classes according to the motif, and the explainer is going to find out the motif in the graph as the explanation. 
Assuming the graph $G^{(e)}$ and $G^{(i)}$ independently follow the distribution $\sP_{\gG^{(e)}}$ and $\sP_{\gG^{(i)}}$ respectively,  denoted as $G^{(e)} \sim \sP_{\gG^{(e)}}$ and $G^{(i)} \sim \sP_{\gG^{(i)}}$,
we randomly sample $G_b = G^{(e)}_b + G_b^{(i)}$ from the data set. Both $G$ and $G_b$ follow the distribution $\sP_{\gG} = \sP_{\gG^{(e)}, \gG^{(i)}}$. We could get our Mixup explanation:
% \begin{equation}
% \begin{split}
%     G^{(\text{mix})} & = \mM + (\mA_b - \mM_b) \\
%                         & \coloneqq G^{(e)} + (G^{\prime} - G^{\prime}^{(e)}) \\
%                         & = G^{(e)}+ G^{\prime}_{i},
% \end{split}
% \end{equation}
\begin{equation}
\begin{aligned}
       G^{(\text{mix})} \coloneqq G^{(e)} + (G_b - G_b^{(e)}) = G^{(e)}+ G_b^{(i)}, 
\end{aligned}
\end{equation}
Then, we have $\sP_{\gG^{(\text{mix})}} = \sP_{\gG^{(e)}} * \sP_{\gG^{(i)}} = \sP_{\gG}$.  It is easy to show that $KL(G, G^{(\text{mix})}) =0$. Thus, we have 
\begin{equation}
    KL(G, G^*) \geq \text{KL}(G, G^{(\text{mix})})
\end{equation}
% \begin{equation}
% \begin{aligned}
% KL(G^{(\text{mix})}, G) & = KL(G^{(e)}+G^{(i)}_b, G^{(e)}+G^{(i)})  \\
%                         & \ll KL(G^{(e)}, G^{(e)}+G^{(i)})  &&\text{\hua{@dongsheng: note 1}} \\
%                         & = KL(G^{(e)}, G) &&\text{\hua{note 2}}
% \end{aligned}
% \end{equation}
\end{proofs}
The theoretical justification shows that our objective function could better estimate the explanation distribution and resolve the distribution shifting issue than the previous approach. In addition, with a safe assumption that $I(G^\Delta, Y | G^*)=0$, as discussed in Eq. (\ref{eq:mixupformula}), we have \ours satisfy the s.t. condition in Eq.~(\ref{eq:gib-ours}). Thus, we can simplify the objective for \ours as:
\begin{equation}
    \argmin_{G^*}  { \quad I(G, G^*)+\alpha \text{CE}(Y, Y^{(\text{mix})})}\\
\end{equation}

\section{Experimental Study}
We conduct comprehensive experimental studies on benchmark datasets to empirically verify the effectiveness of the proposed \ours. Specifically, we aim to answer the following research questions:

% compare with representative and 
% We conducted a series of experiments in this section to compare our method~\ours with other baselines, especially GNNExplainer and PGExplainer. We conducted experiments on six datasets and measured their distribution shifting of them to answer the following research questions:

% \input{tabletexs/expl_auroc.tex}

\begin{table*}[t!]
  \setlength{\tabcolsep}{4.5pt}
  
  \caption{Explanation faithfulness in terms of AUC-ROC on edges under six datasets. The higher, the better. Our mixup approach achieves consistent improvements over backbone GIB-based explanation methods.} 
  % The higher, the better. Our proposed mixup approach can improve existing methods. 
  \label{tab:expl_auroc} 
  \begin{tabular}{c|cccccc}
    \hline
                    & \bashapes                         & \bacom                  & \treec                  & \treeg                     & \bamo                    & \mutag \\
    \hline
    GRAD            & $0.882$                           & $0.750$                       & $0.905$                       & $0.612$                       & $0.717$                       & $0.783$ \\
    ATT             & $0.815$                           & $0.739$                       & $0.824$                       & $0.667$                       & $0.667$                       & $0.765$ \\
    SubgraphX       & $ 0.548$                          & $0.473$                       & $0.617$                       & $0.516$                       & $0.610$                       & $0.529$ \\
    MetaGNN         & $0.851$                           & $0.688$                       & $0.523$                       & $0.628$                       & $0.500$                       & $0.680$   \\
    RG-Explainer     & $0.985$                           & $0.919$                       & $0.787$                       & $0.927$                       & $0.657$                       & $0.873$   \\
    \hline
    GNNExplainer    & $0.884_{\pm 0.002}$               & $0.682_{\pm 0.004}$           & $0.683_{\pm 0.009}$           & $0.379_{\pm 0.001}$           & $0.660_{\pm 0.006}$           & $0.539_{\pm 0.002}$ \\
    + MixUp         & $0.890_{\pm 0.004}$               & $0.788_{\pm 0.006}$           & $0.690_{\pm 0.014}$           & $0.501_{\pm 0.003}$           & $0.869_{\pm 0.004}$           & $0.612_{\pm 0.043}$ \\
    (improvement)   & $0.60\%$                          & $15.5\%$                      & $1.02\%$                      & $32.2\%$                      & $31.7\%$                      & $13.5\%$ \\
    \hline
    PGExplainer     & $0.999_{\pm 0.001}$               & $0.829_{\pm 0.040}$           & $0.762_{\pm 0.014}$           & $0.679_{\pm 0.008}$           & $0.679_{\pm 0.043}$           & $0.843_{\pm 0.084}$ \\
    + MixUp         & $0.999_{\pm 0.001}$               & $0.955_{\pm 0.017}$           & $0.774_{\pm 0.004}$           & $0.712_{\pm 0.000}$           & $0.920_{\pm 0.031}$           & $0.871_{\pm 0.079}$ \\ 
    (improvement)   & $0.00\%$                          & $15.2\%$                      & $1.57\%$                      & $4.86\%$                      & $35.5\%$                      & $3.32\%$ \\
    \hline
  \end{tabular}
  \label{QEtable}
\end{table*}

\begin{itemize}
    \item RQ1: Can the proposed framework outperform the GIB in identifying explanatory substructures for GNNs?
    % Can the proposed approach improve the performance of existing state-of-the-art methods?
    \item RQ2: Is the distribution shifting issue severe in the existing GNN explanation methods? Could the proposed Mixup approach alleviate this issue?
    \item RQ3: How does the proposed approach perform under different hyperparameters?
\end{itemize}

\subsection{Experiment Settings}
\subsubsection{Datasets} 
We focus on analyzing the effects of the distribution shifting problem between the ground truth explanation and the original graphs. Thus, we select six publicly available benchmark datasets with ground truth explanations in our empirical studies~\footnote{All the dataset and codes can be found in \url{https://github.com/jz48/MixupExplainer}}. 
\begin{itemize}
    \item \textbf{\bashapes~}~\cite{ying2019gnnexplainer}: This is a node classification dataset based on a 300-node Barabási-Albert (BA) graph, to which 80 "house" motifs have been randomly attached. The nodes are labeled for use by GNN classifiers, while the edges within the corresponding motif serve as ground truth for explainers. There are four classes in the classification task, with one class indicating nodes in the base graph and the others indicating the relative location of nodes in the motif.
    \item \textbf{\bacom~}~\cite{ying2019gnnexplainer}: This extends the \bashapes dataset to more complex scenarios with eight classes. Two types of motifs are attached to the base graph, with nodes in different motifs having different labels.
    \item \textbf{\treec~}~\cite{ying2019gnnexplainer}: This is a node classification dataset with two classes, with a binary tree serving as the base graph and a 6-node cycle structure as the motif. The labels only indicate if the nodes are in the motifs.
    \item \textbf{\treeg~}~\cite{ying2019gnnexplainer}: This is a node classification dataset created by attaching 80 grid motifs to a single 8-layer balanced binary tree. The labels only indicate if the nodes are in the motifs, and edges within the relative motif are used as ground-truth explanations.
    \item \textbf{\bamo~}~\cite{luo2020parameterized}: This is a graph classification dataset where the label of the graph depends on the type of motif attached to the base graph, which is a BA random graph. The two types of motifs are a 5-node house structure and a 5-node circle structure.
    \item \textbf{\mutag~}~\cite{debnath1991structure}: Unlike other synthetic datasets, \mutag is a real-world molecular dataset commonly used for graph classification explanations. Each graph in \mutag represents a molecule, with nodes representing atoms and edges representing bonds between atoms. The labels for the graphs are based on the chemical functionalities of the corresponding molecules.
\end{itemize}

\subsubsection{Baselines} 
To assess the effectiveness of the proposed framework, we use representative GIB-based explanation methods, GNNExplainer~\cite{ying2019gnnexplainer} and PGExplainer~\cite{luo2020parameterized} as baselines. 
We include these two backbone explainers in our framework \ours and replace the GIB objective with the new proposed mixup objective. The methods are denoted by MixUp-GNNExplainer and MixUp-PGExplainer, respectively.
% The proposed algorithms from Section 4.2 were implemented and integrated into two popular GNN explanation frameworks, GNNExplainer \cite{gnnexp}, and PGExplainer \cite{luo2020parameterized}. 
We also include other types of post-hoc explanation methods for comparison, including GRAD~\cite{ying2019gnnexplainer}, ATT~\cite{velivckovic2017graph}, SubgraphX~\cite{subgraphx}, MetaGNN~\cite{spinelli2022meta}, and RG-Explainer~\cite{rgexp}.

% \dongsheng{Make the names consistent, such as GNNExplainer and GNNExplainer} \jx{got it, we may need to conclude the similarity and difference between these baselines. I think most of them all focus on the substructure without considering the OOD problem}
\begin{itemize}
    \item \textbf{GRAD}~\cite{ying2019gnnexplainer}:  GRAD learns weight vectors of edges by computing gradients of GNN’s objective function.
    \item \textbf{ATT}~\cite{velivckovic2017graph}:  ATT distinguishes the edge attention weights in the input graph with the self-attention layers. Each edge’s importance is obtained by averaging its attention weights across all attention layers.
    \item \textbf{SubgraphX}~\cite{subgraphx}: SubgraphX uses Monte Carlo Tree Search (MCTS) to find out the connected sub-graphs, which could preserve the predictions as explanations.
    \item \textbf{MetaGNN}~\cite{spinelli2022meta} MetaGNN proposes a meta-explainer for improving the level of explainability of a GNN directly at training time by training the GNNs and the explainer in turn. 
    \item \textbf{RG-Explainer}~\cite{rgexp}: RG-Explainer is an RL-enhanced explainer for GNN, which constructs the explanation subgraph by starting from a seed and sequentially adding nodes with an RL agent.
    \item \textbf{GNNExplainer}~\cite{ying2019gnnexplainer}: GNNExplainer is a post-hoc method, which provides explanations for every single instance by learning an edge mask for the edges in the graph. The weight of the edge could be treated as important.
    \item \textbf{PGExplainer}~\cite{luo2020parameterized}: PGExplainer extends GNNExplainer by adopting a deep neural network to parameterize the generation process of explanations, which enables PGExplainer to explain the graphs in a global view. It also generates the substructure graph explanation with the edge importance mask.
\end{itemize}

\subsubsection{Configurations} 
The experiment configurations are set following prior research ~\cite{holdijk2021re}. A three-layer GCN model was trained on 80\% of each dataset's instances as the target model. All explanation methods used the Adam optimizer with a weight decay of 5$e$-4~\cite{kingma2014adam}. The learning rate for GNNExplainer was initialized to 0.01, with 100 training epochs. For PGExplainer, the learning rate was set to 0.003, and the training epoch was 30. The weight of mix-up processing, controlled by $\lambda$, was determined through grid search. Explanations are tested in all instances. While running our approach MixUp-GNNExplainer and MixUp-PGExplainer and comparing them to the original GNNExplainer and PGExplainer, we set them with the same configurations, respectively. Hyperparameters are kept as the default values in other baselines. 

\subsubsection{Evaluation Metrics} 
Due to the existence of gold standard explanations, we follow existing works~\cite{ying2019gnnexplainer, luo2020parameterized, holdijk2021re} and adopt AUC-ROC score on edge importance to evaluate the faithfulness of different methods. 
Other metrics, such as fidelity~\cite{subgraphx}, are not included because the metrics themselves are affected by the distribution shifting problem, making them unsuitable in our setting. 
% \dongsheng{Follow WSDM paper or other related work to describe how to calculate AUC (model as an edge classification problem and compare with the oracle explanations)}

% On benchmarks with predefined explanations available (oracle explanations), we could compute the AUROC score on identified edges, as a well-trained target GNN should follow these explanations. To evaluate the consistency of explanations, we randomly run each method ten times with tuned configurations and report the average score with standard deviation. A smaller standard deviation score indicates stronger consistency. 

To quantitatively measure the distribution shifting between the original graph and the explanation graph, we use \emph{Cosine score} and \emph{Euclidean distance} to measure the distances between the graph embeddings learned by the GNN model. For the Cosine score, the range is $[-1, 1]$, with 1 being the most similar and -1 being the least similar. For the Euclidean distance, the smaller, the better.

\subsection{Quantitative Evaluation (RQ1)}

To answer RQ1, we compare \ours with other baseline methods in terms of the AUC-ROC score. Our approach is evaluated using the weighted vector of the graph generated by the explainers, which serves as the explanation and is compared against the ground truth to calculate the AUC-ROC score. Each experiment is conducted $10$ times with random seeds. We summarize the average performances in Table~\ref{QEtable}.

% summarizes the results of their performances. From the results, we have the following observations:
% The baseline methods we compare against include GRAD, ATT, Gem, RG-Explainer, GNNExplainer, and PGExplainer.
As shown in Table~~\ref{QEtable},  across all six datasets, with both GNNExplainer or PGExplainer as the backbone methods, \ours can consistently and significantly improve the quality of obtained explanations. Specifically, Mixup-GNNExplainer improves the AUC scores by 12.3\%, on average, on the node classification datasets,  and 22.6\% on graph classification tasks. Similarly, MixUp-PGExplainer achieves average improvements of 5.41\% and 19.4\% for node/graph classification tasks, respectively. The comparisons between our \ours and the original counterparts indicate the advantage of the proposed explanation framework. In addition, MixUp-PGExplainer achieves competitive and even state-of-the-art performances compared with other sophisticated baselines, such as reinforcement learning-based RG-Explainer. 

% In addition, in datasets where GNNExplainer and PGExplainer perform worse than GRAD or ATT, like \bamo, applying our mixup approach with the same configurations, the Mixup-GNNExplainer and Mixup-PGExplainer can outperform GRAD than GRAD with $15\%$ and $21\%$ respectively.

\begin{table*}[t!]

  \setlength{\tabcolsep}{4.5pt}
  % $\vv_{\mathcal{G}}$ is the output before the last layer of well-trained GNN model $f$ with the input of the original graph $\mathcal{G}$. $\vv_{\mathcal{G}^{\prime}}$ is the output before $f$'s last layer with the input of the ground truth explanation subgraph $\mathcal{G}^{\prime}$, and $\vv_{\mathcal{G}^\prime_{mix}}$ is with the input of the explanation $\mathcal{G}^\prime_{mix}$ from \ours.
  \caption{The Cosine score and Euclidean distance between the distribution vectors of the original graph $\vh$, explanation subgraph $\vh^*$, and our mixup graph $\vh^{(\text{mix})}$ on different datasets. Large Cosine scores and small Euclidean distances indicate high similarities between representations. The standard deviations of $\text{Avg. Cosine} (\vh, \vh^*)$ and $\text{Avg. Euclidean}(\vh, \vh^*)$ are not included because they are static without random processes.}
  \label{OODtable}
  \begin{tabular}{c|cccccc}
    \hline
                                                                    & \bashapes             & \bacom          & \treec          & \treeg             & \bamo                & \mutag \\
    \hline
    $\text{Avg. Cosine} (\vh, \vh^*)$           & $0.574$               & $0.483$               & $0.962$               & $0.629$               & $0.579$                   & $0.775$ \\
    $\text{ Avg. Cosine} (\vh, \vh^{(\text{mix})})$     & $0.940_{\pm 0.005}$   & $0.644_{\pm 0.006}$   & $0.953_{\pm 0.006}$   & $0.810_{\pm 0.004}$   & $0.901_{\pm 0.000}$       & $0.852_{\pm 0.006}$ \\
    \hline
    $\text{Avg. Euclidean}(\vh, \vh^*)$ & $1.30$                & $1.31$                & $0.213$               & $0.921$               & $1.32$                    & $1.07$ \\
    $\text{Avg. Euclidean}(\vh, \vh^{(\text{mix})})$       & $0.440_{\pm 0.014}$   & $1.10_{\pm 0.010}$    & $0.211_{\pm 0.011}$   & $0.582_{\pm 0.006}$   & $0.587_{\pm 0.001}$       & $0.816_{\pm 0.011}$ \\
    \hline
  \end{tabular} 
\end{table*}

\begin{figure*}
\centering
    \begin{tabular}{cccccc}
    \includegraphics[width=0.3\linewidth]{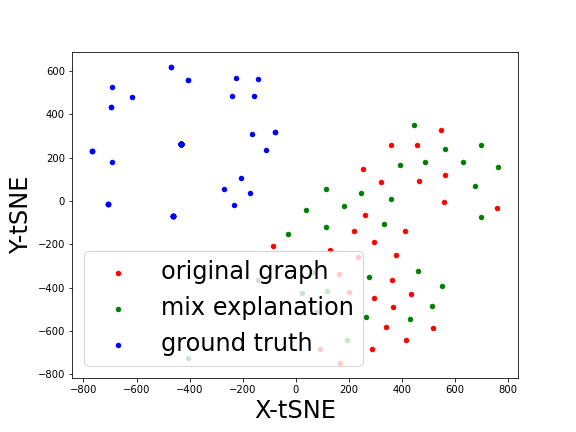}
        & \includegraphics[width=0.3\linewidth]{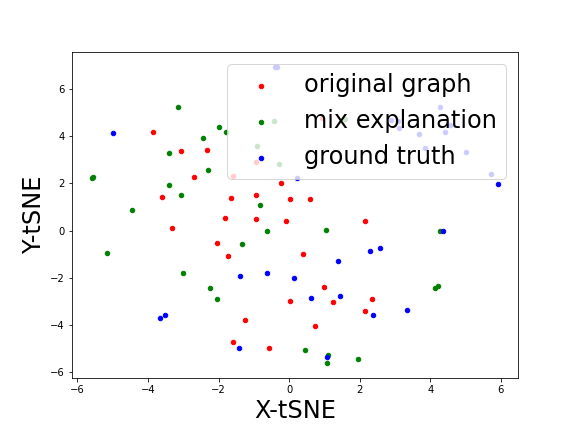} 
        & \includegraphics[width=0.3\linewidth]{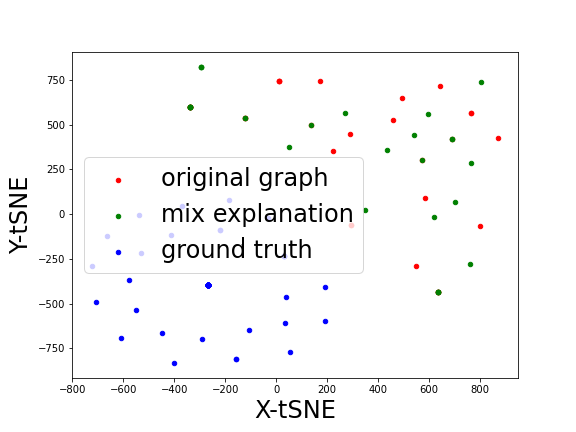} \\ 
    \vspace{-3mm}
        \small{
        \begin{tabular}[c]{@{}c@{}}(a) \bashapes\end{tabular}}& 
        \small{
        \begin{tabular}[c]{@{}c@{}}(b) \bacom\end{tabular}}& 
        \small{
        \begin{tabular}[c]{@{}c@{}}(c) \treec\end{tabular}} \\
        \includegraphics[width=0.3\linewidth]{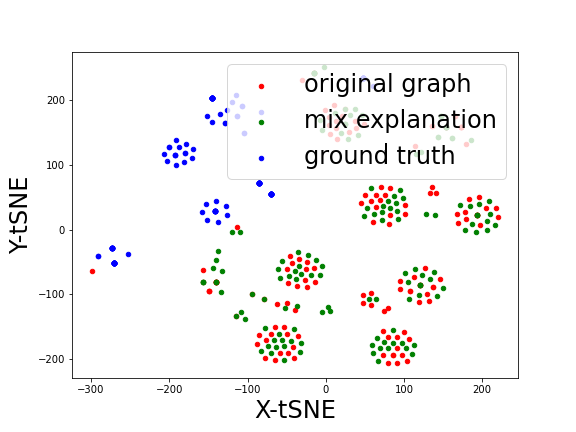}
        & \includegraphics[width=0.3\linewidth]{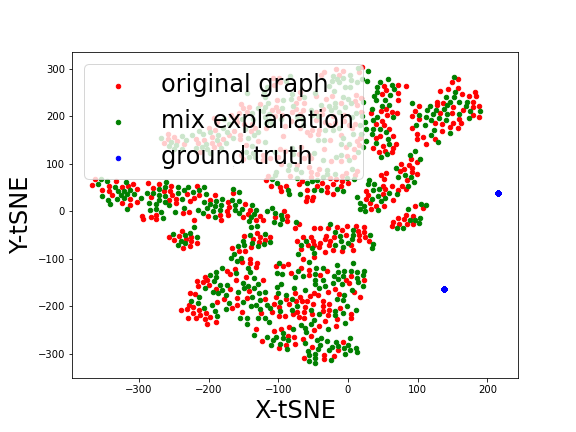} 
        & \includegraphics[width=0.3\linewidth]{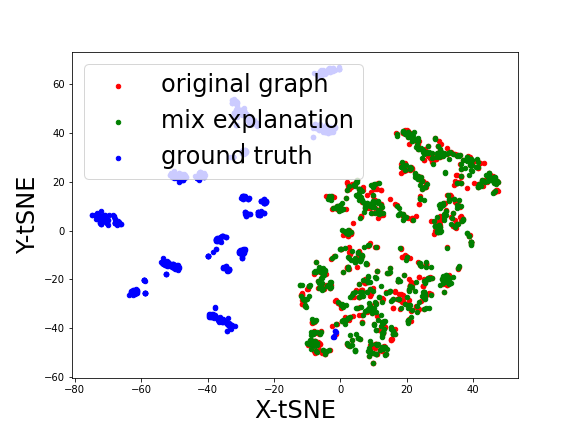} \\
        \small{
        \begin{tabular}[c]{@{}c@{}}(d) \treeg\end{tabular}}& 
        \small{
        \begin{tabular}[c]{@{}c@{}}(e) \bamo\end{tabular}}& 
        \small{
        \begin{tabular}[c]{@{}c@{}}(f) \mutag\end{tabular}}
    \end{tabular} \\
    \vspace{-2mm}
    \caption{Visualizations of the distribution shifting issue with t-SNE on six datasets. The points are generated with the output before the last layer of the model to be explained $f$, which is then plotted with t-SNE. The red points mean original graphs $G$, the blue points mean substructure explanations $G^*$, and the green points mean mixup explanations $G^{(mix)}$. Green dots align well with red dots, while blue dots shift away from red dots.}
    \label{fig:ood}
\end{figure*}

\subsection{Alleviating Distribution Shifts (RQ2)}
\label{sec:rq2}
In the previous section, we showed that our MixUp approach outperforms existing explanation methods in terms of AUC-ROC. In this section, we show the existence of the distribution shifting issue and show our proposed mixup approach alleviates this issue and improves the performance in explanation w.r.t. AUC. 

\stitle{Visualizing Distributing Shifting.}
In this section, we show the existence of the distribution shifting issue by visualizing the distribution vector (the output of the last layer in a well-trained GNN model $f$) for the original graph $G$, the explanation from \ours $G^{(\text{mix})}$, and the ground truth explanation $G^*$ with t-Distributed Stochastic Neighbor Embedding(t-SNE)~\cite{van2008visualizingtsne}. 
To calculate distribution vectors, we use the output of the last GNN layer in $f$ as the representation vector $\vh$ for the original graph. Ground truth explanations $G^*$ and the mixup graph from \ours, $G^{(\text{mix})}$ are also fed into the model to achieve corresponding representations, denoted by $\vh^*$, $\vh^{(\text{mix})}$, respectively.
The visualization results can be found in Figure~\ref{fig:ood}. The red points represent the vectors from original graphs $\mathcal{G}$; the blue points represent vectors from substructure explanations $\mathcal{G}^\prime$, and the green points represent the vectors from the mixup explanations $\mathcal{G}^\prime_{mix}$. Note that for \bamo, while there are multiple graphs in the dataset, with only two kinds of motifs as explanations, t-SNE only shows two blue points, which are actually multiple overlapping blue points. From Figure~\ref{fig:ood}, we have the following observations:
\\~$\bullet$ The blue points shift away from the red points in most datasets, including both synthetic and real-world datasets. It means that the distribution shifting issue exists in most cases, where most existing work overlooked this issue.
\\~$\bullet$ The green points are inseparable from the red points in most datasets. It means that the explanation from \ours aligns well with the original graph's distribution, which indicates our mixup approach's effectiveness in alleviating the distribution shifting issue.
\\~$\bullet$ The shifting between blue points and red points is more obvious in the \mutag dataset, where the green points generated by \ours still align well with the red points. This shows our method with mixup works well not only in synthetic datasets but also in the real-world dataset.

\begin{figure}[thb]
    \centering
    \includegraphics[width=0.98\linewidth]{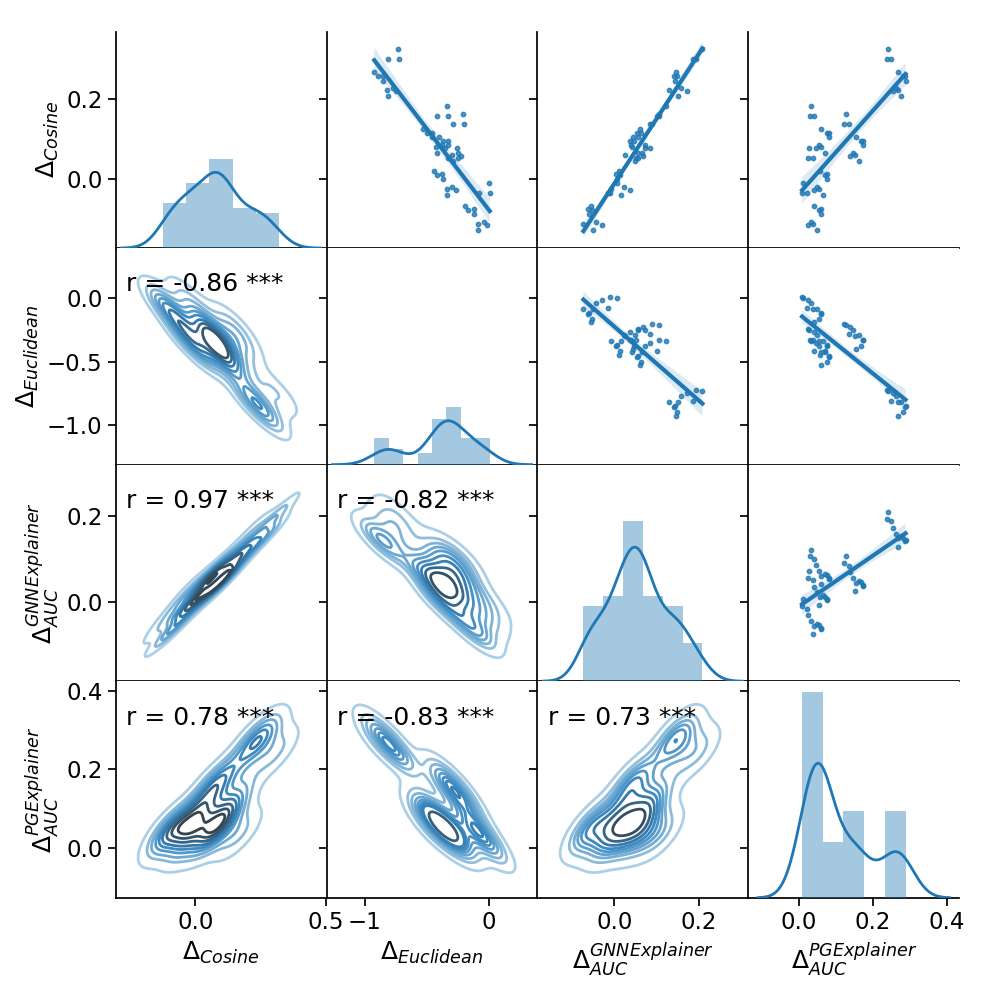}
    \caption{Correlation between improvements of AUC-ROC scores in explanation performance and the improvements of distribution distances on different datasets. The value of $r$ indicates the Pearson correlation coefficient, and the values with $*$ indicate statistical significance for correlation, where $^{***}$ indicates the p-value for testing non-correlation $p \leq 0.001$.}
    \label{fig:corr}
\end{figure}

\stitle{Measuring Distances.} In this section, we quantitatively assess the distribution shifting issue by measuring the distances between the distribution vector $\vh$ from the original graphs and the explanation subgraphs $\vh^*$ and $\vh^{(\text{mix})}$. %Given an input graph $G$ and a trained GNN model $f$, we use the output of the last GNN layer as the representation vector $\vh$ for the original graph. Ground truth explanations $G^*$ and the mixup graph from \ours, $G^{(\text{mix})}$ are also fed into the model to achieve corresponding representations, denoted by $\vh^*$, $\vh^{(\text{mix})}$, respectively. 
We report the averaged Cosine score and the Euclidean distance between different types of representation vectors in Table \ref{OODtable}. From the results, we can see that, on average, $\vh^{(\text{mix})}$ has a higher Cosine score and a smaller Euclidean distance with $\vh$ than $\vh^*$, indicating more similarity of distribution between $G^{(\text{mix})}$ and $G$ than that between $G^*$ and $G$. 
The smaller distances between representation vectors demonstrate that our Mixup approach can effectively alleviate the distribution shifting problem caused by the inductive bias in the prediction model $f$. 
As $G^{(\text{mix})}$ better estimates the distribution of the original graphs, \ours can consistently improve the performance of existing explainers. 

\stitle{Correlation with Performance Improvements.}
We quantitatively evaluate the correlation between the improvements of AUC-ROC scores of \ours over basic counterparts and the improvements in distances with our mixup approach. We calculate the improvements of AUC-ROC scores from GNNExplainer and PGExplainer over GNNExplainer and PGExplainer without mixup (denoted as $\Delta_{\text{AUC}}^{\text{GNNExplainer}}$ and $\Delta_{\text{AUC}}^{\text{PGExplainer}}$, respectively). The improvements of average $\text{Cosine}(\vh, \vh^{(\text{mix})})$ over average $\text{Cosine}(\vh, \vh^{*})$ is denoted by $\Delta_{\text{Cosine}}$, and the improvements on Euclidean distance is 
% \vspace{+6.5mm}
$\Delta_{\text{Euclidean}}$. Figure~\ref{fig:corr} shows the correlation between $\Delta_{\text{AUC}}^{\text{GNNExplainer}}$,  $\Delta_{\text{AUC}}^{\text{PGExplainer}}$, $\Delta_{\text{Cosine}}$, and $\Delta_{\text{Euclidean}}$. We can see that all these four improvements strongly correlated to each other with statistical significance, indicating the improvements achieved by \ours in explanations accuracy own to the successful alleviation of the distribution shifting issue. 

\begin{figure}
\centering
    \begin{tabular}{cc}
    \includegraphics[width=0.45\linewidth]{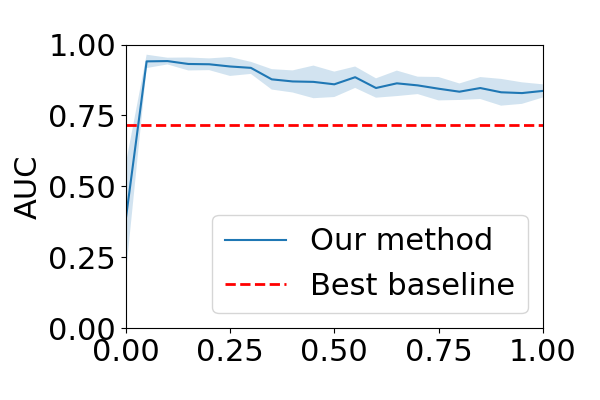}
        & \includegraphics[width=0.48\linewidth]{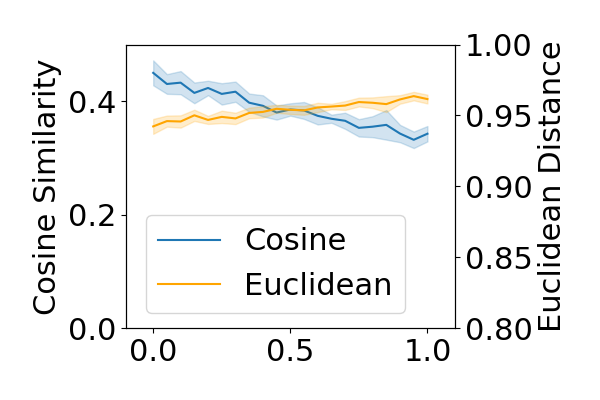} \\
        \small{
        \begin{tabular}[c]{@{}c@{}}(a) AUC for different $\lambda$\end{tabular}}& 
        \small{
        \begin{tabular}[c]{@{}c@{}}(b) Distances for different $\lambda$\end{tabular}}
    \end{tabular} \\
    \caption{Hyper parameter analysis of $\lambda$ on \bamo with Mixup-PGExplainer. (a) The performance of explanation w.r.t AUC. The blue line represents the mean AUC score with standard deviations over ten runs with different random seeds on each $\lambda$ value. The red line represents the performance of the baseline PGExplainer. (b) The distances between $\vh$ and $\vh^{(\text{mix})}$ for different $\lambda$. The blue and yellow lines represent the mean of the Cosine score and Euclidean distance with standard deviations, respectively.}
    \label{fig:tunelambda}
\end{figure}

% \vspace{+10mm}
\subsection{Parameter Study (RQ3)}
% We tune and study the hyperparameters in this section, which include $\lambda$ in formula \ref{mixup_eq} and $\eta$ as the number of edges to connect the explanations with the base graph, on dataset \bamo. We tune the hyperparameter $\lambda \in [0, 1]$ to figure out the best $\lambda$ for our new mixup approach and verify the robustness of the performance of the approach. As shown in figure \ref{fig:tunelambda}, the explainer achieved the best performance with $\lambda = 0.1$. And with all $\lambda$ from 0.05 to 1.0, the explainer performance the baseline 0.679, which means our approach is robust with the hyperparameter $\lambda$. While $\lambda = 0$, the formula degenerates to a trivial solution and could not benefit the explainer. We also tune the hyperparameter $\eta \ in [1, 20]$, which means we randomly sampled $[1, 20]$ edges to connect $G_1$ and $G_2$, where $G_1$ contains the explanation and $G_2$ contains the base structure. As shown in figure \ref{fig:tuneyita}, while we used only 1 edge to connect two graphs, the explainer achieved the best performance. And it obeys the nature of the \bamo dataset that the motif is added to the base structure with 1 edge. Also, it could be observed that our approach is robust with other numbers of connecting edges. With all settings of the $\eta$, our approach out performances the baseline 0.679.

\begin{figure}
\centering
    \begin{tabular}{cc}
    \includegraphics[width=0.45\linewidth]{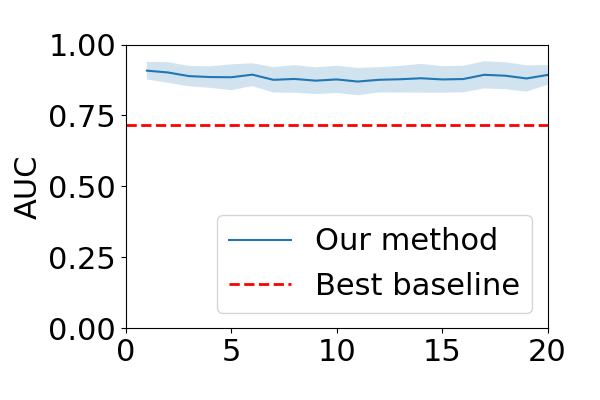}
        & \includegraphics[width=0.48\linewidth]{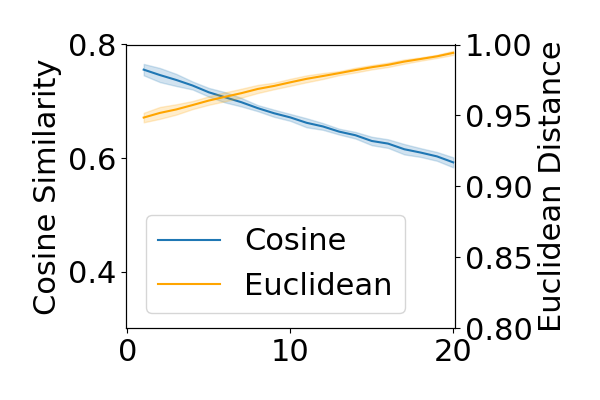} \\
        \small{
        \begin{tabular}[c]{@{}c@{}}(a) AUC for different $\eta$\end{tabular}}& 
        \small{
        \begin{tabular}[c]{@{}c@{}}(b) Distances for different $\eta$\end{tabular}}
    \end{tabular} \\
    \caption{Hyperparameter analysis of $\eta$  on \bamo with Mixup-PGExplainer. (a) The performance of explanation w.r.t AUC. (b) The distances between $\vh$ and $\vh^{(\text{mix})}$ for different $\eta$.}
    \label{fig:tuneyita}
\end{figure}

In this section, we investigate the hyperparameters of our approach, which include $\lambda$ and $\eta$, on the \bamo dataset. The hyperparameter $\lambda$ controls the weight on the original graph during the mixup process. We find the optimal value of $\lambda$ by tuning it within the $[0, 1]$ range. Note that, with $\lambda = 0$, Eq.~(\ref{eq:mixup-ours-lmd}) is trivial and doesn't help explain $G_a$ with only $\mA_b$. The experimental results can be found in Figure~\ref{fig:tunelambda}. We can see that the best performance is achieved with $\lambda = 0.1$ and that the approach consistently outperforms the best performance from baselines with $\lambda \in [0.05, 1]$.
The hyperparameter $\eta$ is the number of cross-graph edges during mixup, indicating the connectivity between label-dependent explanations and label-independent subgraphs. We tune it within the $[1, 20]$ range on the \bamo dataset. The results in Figure~\ref{fig:tuneyita} show that the best performance is achieved with $\eta = 1$.
%, which is also in line with the nature of the \bamo dataset~\hua{why???, }\jx{done}. Because in the \bamo dataset, the motif is connected to the base graph with one edge. 
With different $\eta$, our approach shows stable and consistently better performance than the best baseline.

\section{Conclusion}

In this work, we study the distribution shifting problem to obtain robust explanations for GNNs, which is largely neglected by the existing GIB-based post-hoc instance-level explanation framework. With a close analysis of the explanation methods of GNNs, we emphasize the possible distribution shifting issue induced by the existing framework. We propose a simple yet effective approach to address the distribution shifting issue by mixing up the explanation with a randomly sampled base graph structure. The designed algorithms can be incorporated into existing methods with no effort. Experiments validate its effectiveness, and further theoretical analysis shows that it is more effective in alleviating the distribution shifting issue in graph explanation. In the future, we will seek more robust explanations. Increased robustness indicates stronger generality and could provide better class-level interpretation at the same time.

\section*{ACKNOWLEDGMENTS}

The work was partially supported by NSF award \#2153311. The views and conclusions contained in this paper are those of the authors and should not be interpreted as representing any funding agencies.


\bibliographystyle{ACM-Reference-Format}
\balance
\bibliography{sample-base}

\appendix

\section{APPENDIX}

\subsection{Graph Mixup Algorithm}

\renewcommand{\algorithmicrequire}{\textbf{Input:}}
\renewcommand{\algorithmicensure}{\textbf{Output:}}

\begin{algorithm}
	\caption{Graph Mixup Algorithm} 
	\begin{algorithmic}[1]
        \Require Graph $G_a = (\mX_a, \mA_a)$, a set of graphs $\gG$, the number of random connections $\eta$, explanation model $g$. %Each graph $G$ has an adjacency matrix $\mA$ and node feature matrix $\mX$.
        \Ensure Graph $G^{(\text{mix})}$.
        \State Randomly sample a graph $G_b = (\mA_b, \mX_b)$ from $\gG$
        \State Generate mask matrix $\mM_a=g(G_a)$
        \State Generate mask matrix $\mM_b=g(G_b)$
        % \NoNumber{$\triangleright$\ Align adjacency matrices}
        % \State Extend $\mA_b \gets \mA_b + number\ of\ nodes\ in\ G_a$  % (done fixing compile issue)
        \State Sample $\eta$ random connections between $G_a$ and $G_b$ as $\mA_c$
        \State Mixup adjacency matrix 
        $\mA^{(\text{mix})}$ with Eq.~(\ref{eq:align-adj})      
        \State Mixup edge mask $\mM^{(\text{mix})}$ with Eq.~(\ref{eq:align-mask})
        % \NoNumber{$\triangleright$\ Mixup step}
        %\State $\mM_{mix} \gets \lambda \mM_a + (\mA_b - \lambda \mM_b)$
		\State Mixup node features $\mX^{(\text{mix})} =[\mX_a; \mX_b]$
        \State \Return $G^{(\text{mix})}=(\mX^{(\text{mix})},\mM^{(\text{mix})} \odot \mA^{(\text{mix})})$
    \end{algorithmic} 
    \label{alg1}
\end{algorithm}

\end{document}